\documentclass[referee, pdflatex,sn-mathphys-num]{sn-jnl}

\usepackage{graphicx}%
\usepackage{multirow}%
\usepackage{amsmath,amssymb,amsfonts}%
\usepackage{amsthm}%
\usepackage{mathrsfs}%
\usepackage[title]{appendix}%
\usepackage{xcolor}%
\usepackage{textcomp}%
\usepackage{manyfoot}%
\usepackage{booktabs}%
\usepackage{algorithm}%
\usepackage{algorithmicx}%
\usepackage{algpseudocode}%
\usepackage{listings}%
\usepackage{booktabs} 

\begin{document}

\title[STT]{Fast and Memory-efficient Non-line-of-sight Imaging with Quasi-Fresnel Transform}


\author[1]{\fnm{Yijun} \sur{Wei}}\email{weiyj21@mails.tsinghua.edu.cn}
\equalcont{These authors contributed equally to this work.}

\author[2]{\fnm{Jianyu} \sur{Wang}}\email{jy-wang19@mails.tsinghua.edu.cn}
\equalcont{These authors contributed equally to this work.}

\author[3,4]{\fnm{Leping} \sur{Xiao}}\email{xlp21@mails.tsinghua.edu.cn}

\author*[2,5]{\fnm{Zuoqiang} \sur{Shi}}\email{zqshi@tsinghua.edu.cn (Tel: +86(10)62796455)}

\author*[3,4]{\fnm{Xing} \sur{Fu}}\email{fuxing@tsinghua.edu.cn (Tel: +86(10)62796820)}

\author*[2,5]{\fnm{Lingyun} \sur{Qiu}}\email{lyqiu@tsinghua.edu.cn (Tel: +86(10)62790121)}

\affil[1]{\orgdiv{Zhili College}, \orgname{Tsinghua University}, \orgaddress{\city{Beijing}, \postcode{100084}, \country{China}}}

\affil[2]{\orgdiv{Yau Mathematical Sciences Center}, \orgname{Tsinghua University},\orgaddress{\city{Beijing}, \postcode{100084}, \country{China}}}

\affil[3]{\orgdiv{Department of Precision Instrument}, \orgname{Tsinghua University}, \orgaddress{\city{Beijing}, \postcode{100084}, \country{China}}}

\affil[4]{\orgname{State Key Laboratory of Precision Space-time Information Sensing Technology}, \orgaddress{\city{Beijing}, \postcode{100084}, \country{China}}}

\affil[5]{\orgname{Yanqi Lake Beijing Institute of Mathematical Sciences and Applications},\orgaddress{\city{Beijing}, \postcode{101408}, \country{China}}}

\maketitle

\newpage

\section*{Abstract}
    Non-line-of-sight (NLOS) imaging seeks to reconstruct hidden objects by analyzing reflections from intermediary surfaces. Existing methods typically model both the measurement data and the hidden scene in three dimensions, overlooking the inherently two-dimensional nature of most hidden objects. This oversight leads to high computational costs and substantial memory consumption, limiting practical applications and making real-time, high-resolution NLOS imaging on lightweight devices challenging. In this paper, we introduce a novel approach that represents the hidden scene using two-dimensional functions and employs a Quasi-Fresnel transform to establish a direct inversion formula between the measurement data and the hidden scene. This transformation leverages the two-dimensional characteristics of the problem to significantly reduce computational complexity and memory requirements. Our algorithm efficiently performs fast transformations between these two-dimensional aggregated data, enabling rapid reconstruction of hidden objects with minimal memory usage. Compared to existing methods, our approach reduces runtime and memory demands by several orders of magnitude while maintaining imaging quality. The substantial reduction in memory usage not only enhances computational efficiency but also enables NLOS imaging on lightweight devices such as mobile and embedded systems. We anticipate that this method will facilitate real-time, high-resolution NLOS imaging and broaden its applicability across a wider range of platforms.

\noindent\textbf{Keywords:} Non-line-of-sight imaging,  Inverseproblem, Quasi-Fresnel Transform, Efficient imaging algorithm

\section*{Introduction}
\begin{figure*}[t]
\centering
\includegraphics[width=0.8\textwidth]{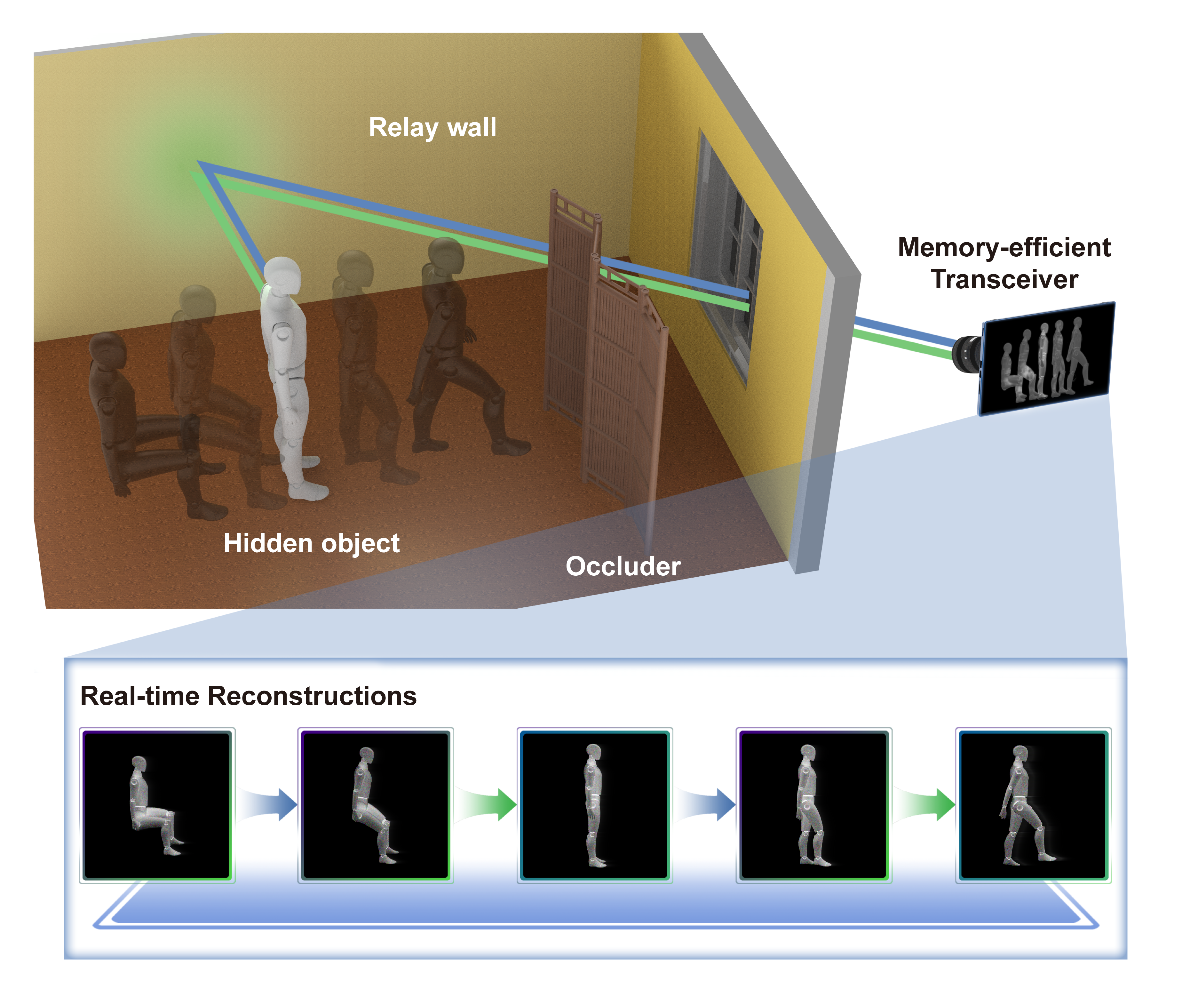}
\caption{Typical setup of NLOS scene. An optical system is employed to measure the reflected light of the hidden object from the relay wall. The proposed algorithm advances the field by enabling real-time reconstruction of the hidden scene, which is crucial for NLOS imaging technology. }
\label{fig: setup}
\end{figure*}

Transient imaging, powered by single-photon avalanche diodes (SPADs), has opened new frontiers in optical imaging, with one of the most compelling applications being non-line-of-sight (NLOS) imaging. Recently, NLOS imaging has attracted significant attention due to its wide range of potential applications, from surveillance and autonomous navigation to medical imaging and archaeology \cite{faccio2020non,rapp2020advances}.
As shown in Fig. \ref{fig: setup}, in a typical NLOS setup, the emitted light is directed to a relay wall, bounces to illuminate the hidden scene, and then reflects back to the relay wall before reaching the sensor. Utilizing the two-dimensional mapping established by the proposed method, hidden geometry can be efficiently reconstructed on handheld devices, providing substantial improvements in both runtime and memory efficiency.

\begin{figure*}[t]
\centering
\includegraphics[width=0.9\textwidth]{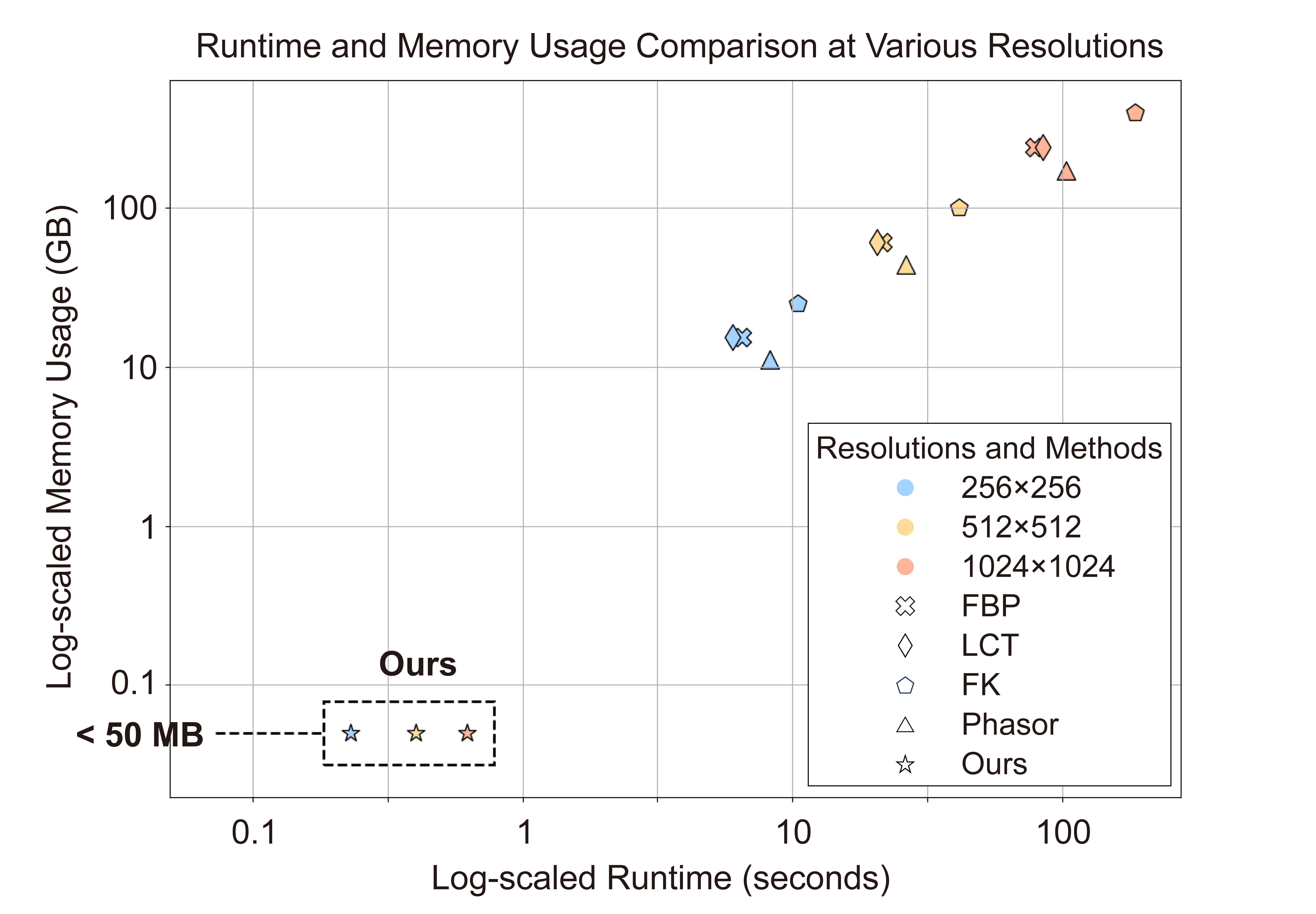}
\caption{Runtime and memory usage comparison between different methods. Compared to previous methods, the proposed algorithm requires significantly less runtime and memory to reconstruct the hidden scene through a two-dimensional relationship between the measurement and the hidden scene.}
\label{fig: runtime comparison}
\end{figure*}

NLOS imaging problem has become an emerging field since 2009 \cite{kirmani2009looking}. Several methods have been developed to enhance the practicality of this technique. The reconstruction algorithms can be categorized into three main categories: imaging algorithms \cite{velten2012recovering,feng2020improving,xu2022fast,o2018confocal,lindell2019wave,liu2019non,liu2020phasor,yu2023enhancing,mu2022physics,grau2020deep}, which directly yield reconstructions; iterative algorithms \cite{plack2023fast,young2020non,liu2021non,heide2019non,iseringhausen2020non,tsai2019beyond,liu2023non,liu2023few,wu2021non}, which iteratively improve reconstruction quality; and learning-based algorithms \cite{metzler2020deep,chen2019steady,isogawa2020optical,shen2021non,chen2020learned,li2023nlost,li2024deep}, which utilize neural networks. 
Despite these advances, producing real-time high-resolution NLOS videos remains challenging, as existing methods cannot compute high-resolution results in fractions of a second; although they can achieve hundreds of FPS at very low resolutions (e.g., $32 \times 32$), they cannot simultaneously deliver high spatial resolution and real-time performance. Additionally, most existing methods are memory-intensive, making them impractical for lightweight devices with limited computing power, such as embedded systems. This is because they use three-dimensional volumes for reconstruction, overlooking the two-dimensional characteristics of the hidden object, which increases the memory and computational complexity. 

To address these challenges, we introduce a novel approach that exploits the two-dimensional nature of the hidden scene through the Quasi-Fresnel transform. The term “Quasi-Fresnel” highlights that our approach, while conceptually related to the traditional Fresnel integral in wave optics, is specifically adapted to the unique structure of the NLOS imaging problem.
The proposed Quasi-Fresnel transform differs from the traditional Fresnel integral by focusing on key wavefront properties, specifically how light interacts with intermediary surfaces and hidden objects in the NLOS context. By applying this transform, we establish a two-dimensional relationship between the measurement data and the hidden scene, effectively reducing the dimensionality of the problem and enabling more efficient computation. When the measurement has $N \times N $ scanning points on the relay wall, the computational and memory complexity of the proposed method are $O(N^2 \log N)$ and $O(N^2)$. This reduction in complexity enables real-time, high-resolution reconstructions on lightweight devices, a feat unattainable with traditional volumetric methods.

As illustrated in Fig. \ref{fig: runtime comparison}, the proposed algorithm outperforms previous imaging algorithms in terms of both runtime and memory usage. Even when reconstructing a hidden scene with a size of $1024 \times 1024$, the memory usage of the proposed method is still less than 50 MB, which is thousands of times less than previous methods. In addition, the runtime of the proposed method is under 1 s, also outperforming previous methods by hundreds of times.
This comparison uses simulated data with a kylin as the hidden object, with further details provided in the Method section. Corresponding reconstruction results are provided in the Supplementary Information.
Moreover, reconstruction results on public datasets demonstrate that our method delivers much more efficient reconstruction while maintaining imaging quality.

Our method is efficient for the confocal scenario, where the illumination point coincides with the detection point, while other NLOS scenarios can also be converted into the confocal case for processing \cite{lindell2019wave,liu2023non}. 
The proposed algorithm not only enables real-time, high-resolution NLOS imaging but also offers insights for future hardware development.

\section*{Results}
\subsection*{Efficient NLOS Reconstruction via the Quasi-Fresnel Transform}

\begin{figure*}[t]
\centering
\includegraphics[width=0.9\textwidth]{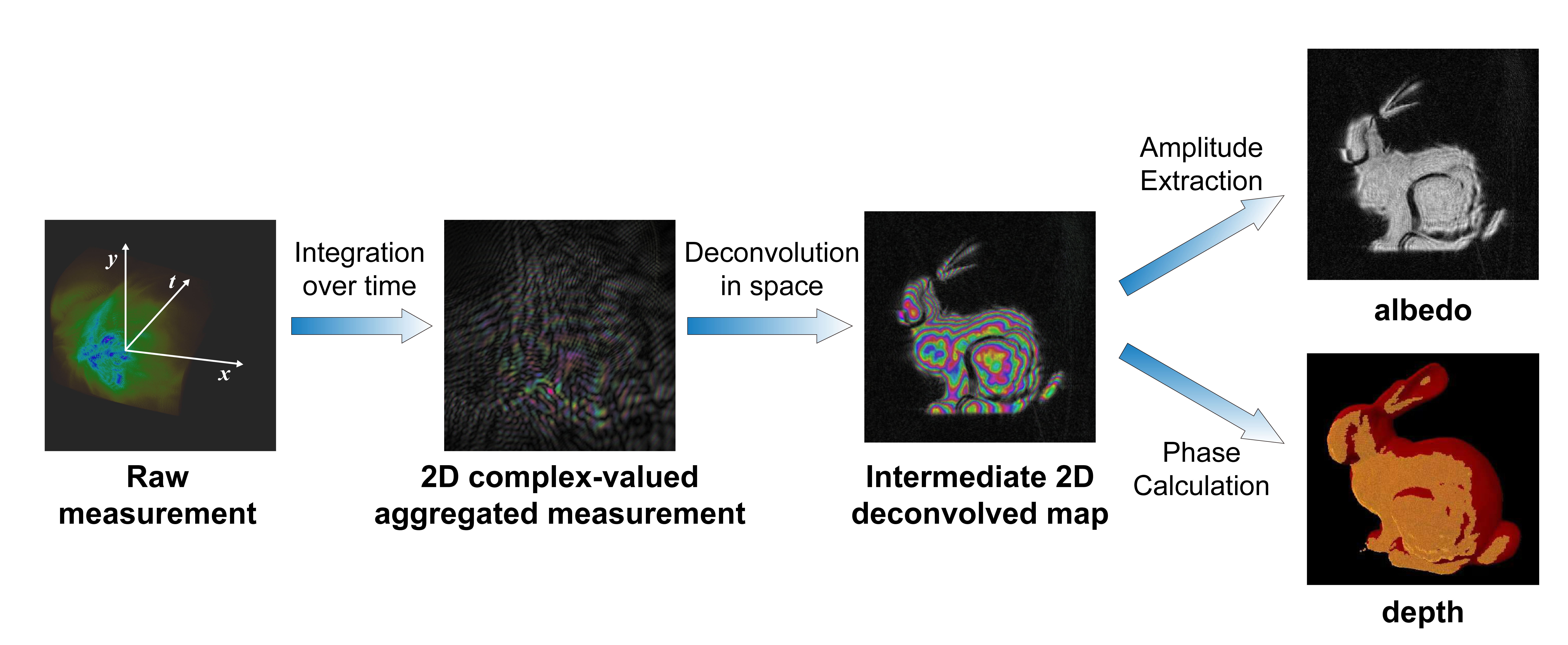}
\caption{An illustration of the reconstruction pipeline of the proposed algorithm. The three-dimensional raw measurement is first integrated over time to reduce its dimension, resulting in a two-dimensional complex-valued aggregated measurement. This is followed by a spatial deconvolution step, from which the albedo and depth are extracted. The complex-valued data is visualized in a figure where hue represents the phase, and saturation and value represent the amplitude. }
\label{fig: pipeline}
\end{figure*}

NLOS imaging is traditionally constrained by the computational complexity of processing three-dimensional data to reconstruct hidden scenes. Existing methods typically rely on volumetric models and three-dimensional fast Fourier transform (FFT), resulting in substantial computational overhead. 

To overcome these limitations, we propose a transformative approach that utilizes the inherent two-dimensional nature of the hidden scene. By representing the volumetric albedo $f(\boldsymbol{y})$ of the hidden scene as $f(\boldsymbol{x},z)=a(\boldsymbol{x}) \delta(z-d(\boldsymbol{x})) , \,\boldsymbol{x}\in\mathbb{R}^2$, where $a(\boldsymbol{x})$ and $d(\boldsymbol{x})$ represent the albedo and depth maps, respectively, we shift from a three-dimensional reconstruction approach to a more efficient two-dimensional method. This key insight allows us to bypass the high complexity of traditional methods.

At the core of our approach is the introduction of the Quasi-Fresnel transform. This transform maps the three-dimensional volumetric albedo, $f(\boldsymbol{y})$, to a two-dimensional function $\phi(\boldsymbol{x}; s)$. Specifically, we define $\phi(\boldsymbol{x}; s)$ as:

\begin{equation}
\label{eq: def phi}
      \phi(\boldsymbol{x}; s) = \int_{\mathbb{R}^3} e^{-i \frac{\| (\boldsymbol{x},0) - \boldsymbol{y}\|^2}{4s^2}} f(\boldsymbol{y}) d \boldsymbol{y},
\end{equation}
where $i=\sqrt{-1}$ and $s>0$ is a fixed parameter selected during implementation, controlling the spread of the wavefront in space. This formulation defines the Quasi-Fresnel transform and is central to our method for efficient NLOS imaging.

In NLOS, the measurement is modeled using the three-point transport model \cite{veach1995optimally}. Without considering multiple reflections or occlusions, the forward model can be described as\cite{o2018confocal}:
\begin{equation}
    \label{eq: forward model}
    \tau(\boldsymbol{x}, t) = \frac{1}{(\frac{ct}{2})^k} \int_{\partial B((\boldsymbol{x}, 0), \frac{ct}{2})} f(\boldsymbol{y}) \, dA(\boldsymbol{y}),
\end{equation}
where $\tau(\boldsymbol{x}, t)$ represents the signal measured at the point $(\boldsymbol{x},0) \in \mathbb{R}^3$ on the relay wall $\{z=0\}$ and time $t$, $c$ is the speed of light, and $k$ is $4$ for isotropic scattering or $2$ for retroreflective materials. $\partial B((\boldsymbol{x}, 0), \frac{ct}{2})$ denotes the sphere centered at $(\boldsymbol{x},0)$ with radius $\frac{ct}{2}$, and $A(\boldsymbol{y})$ is the corresponding area measure. 
Building on this forward model, $\phi(\boldsymbol{x}; s)$ can also be understood as an aggregation of the measured data $\tau(\boldsymbol{x}, t)$ along the time dimension. It is rewritten as:
\begin{equation}
    \label{eq: blur 3d}
        \phi(\boldsymbol{x}; s) =\frac{c^{k+1}}{2^{k+1}}\int_0^{\infty} e^{- i \frac{c^2 t^2}{16 s^2}  } t^k \tau(\boldsymbol{x}, t)  d t.
    \end{equation}
By applying this transform with specific values of $s$, we reduce the dimensionality of the measurement data, significantly enhancing the efficiency of data processing and storage. 

Next, we define $\psi(\boldsymbol{x}; s)$, which arises due to the separability of the kernel in the definition of $\phi(\boldsymbol{x}; s)$. Specifically, $\psi(\boldsymbol{x}; s)$ represents the result of the one-dimensional Quasi-Fresnel transform of $f$ along the depth axis:
\begin{equation}
    \label{eq: define psi}
    \psi(\boldsymbol{x}; s)=a(\boldsymbol{x}) e^{- i \frac{d(\boldsymbol{x})^2}{4 s^2} }.
\end{equation}
This allows us to directly extract the albedo and depth information as $| \psi |$ and $\left(-2i s^3 \psi^{-1} \frac{\partial \psi}{\partial s}\right)^{1 / 2}$, respectively. The inversion process, shown in the following equation, further connects the aggregated measurement $\phi(\boldsymbol{x}; s)$ and the modulated albedo $\psi(\boldsymbol{x}; s)$:
\begin{equation}
    \label{eq: relationship between functions}
       \psi(\boldsymbol{x}; s) =  \frac{1}{16 \pi^2 s^4} \int_{\mathbb{R}^2} e^{ i \frac{\| \boldsymbol{\tilde{x}} - \boldsymbol{x}\|^2 }{4 s^2} } \phi(\boldsymbol{\tilde{x}}; s) d \boldsymbol{\tilde{x}}.  
\end{equation}
This formulation enables efficient reconstruction of the hidden scene using a two-dimensional FFT, significantly reducing both memory usage and computational complexity. By applying the Quasi-Fresnel transform, we condense the three-dimensional measurement into a two-dimensional representation through optimized parameter choices for $s$. Our method offers a more computationally feasible solution for NLOS imaging while preserving high image quality.

Figure \ref{fig: pipeline} illustrates the subprocesses of the proposed algorithm. The simulated measurement of the Stanford bunny is provided by the Zaragoza dataset\cite{galindo19-NLOSDataset}. The proposed algorithm consists of three main steps: (I) integrating over the time dimension, which converts the reconstruction problem into a two-dimensional deconvolution problem; (II) solving the deconvolution problem in the spatial dimensions; and (III) obtaining the final reconstructions from the recovered $\psi$. These steps are efficient in terms of both runtime and memory usage. If the first step is executed on a computer, then the proposed method's computational and memory complexities are both $O(N^3)$. However, the integration step can be efficiently implemented on hardware, reducing the computational and memory complexities to $O(N^2 \log N)$ and $O(N^2)$, respectively. A more detailed discussion of the complexity is provided in the Method section. This advancement significantly accelerates the development of real-time NLOS imaging.  

\subsection*{Results on public datasets}

\begin{figure*}[t]
\centering
\includegraphics[width=1.0\textwidth]{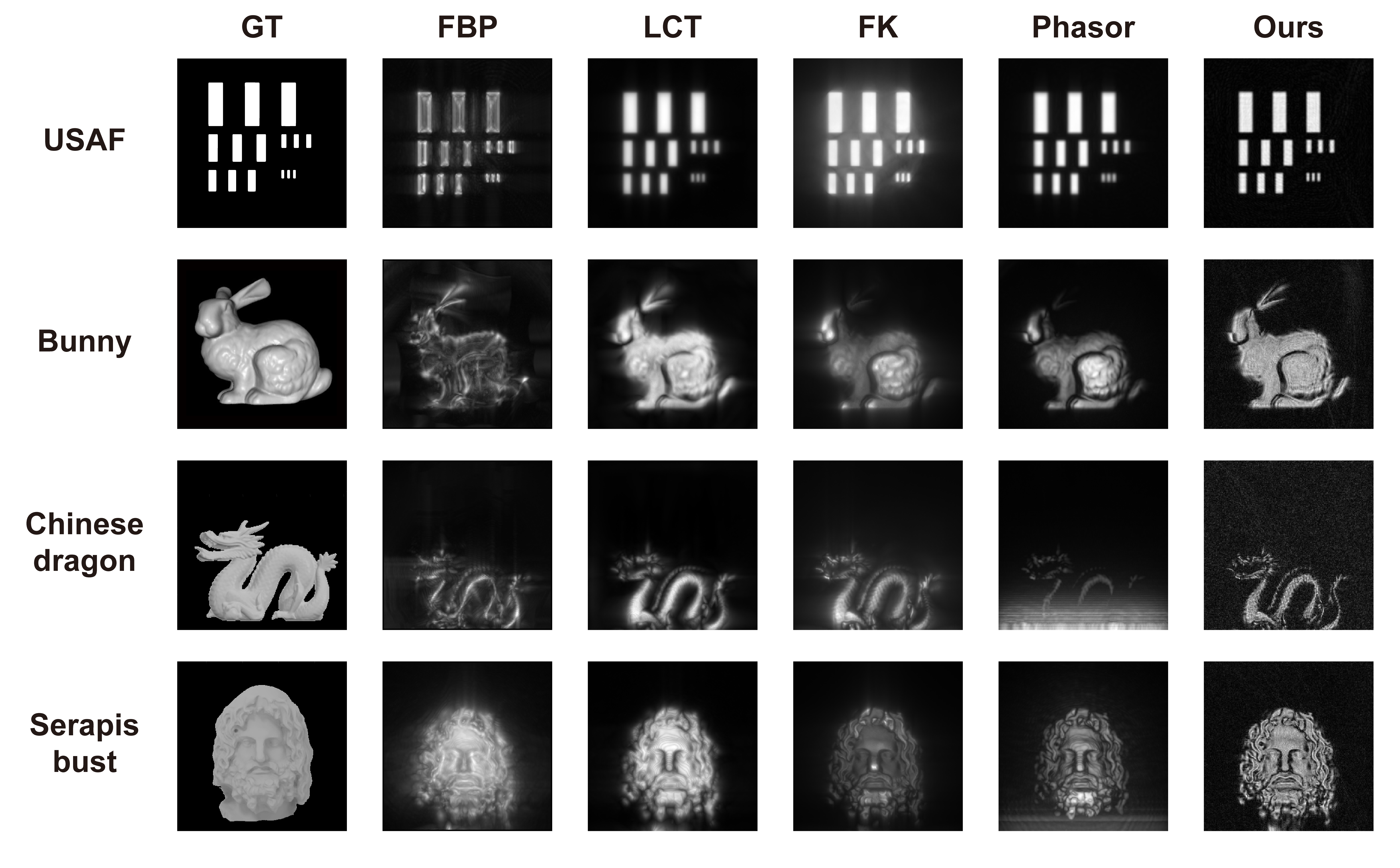}
\caption{Reconstructions of the Zaragoza dataset. ``GT" denotes the ground truth of the hidden object. Compared to previous methods, the proposed method reconstructs the hidden object more faithfully.}
\label{fig: results zaragoza}
\end{figure*}

\begin{figure*}[t]
\centering
\includegraphics[width=1.0\textwidth]{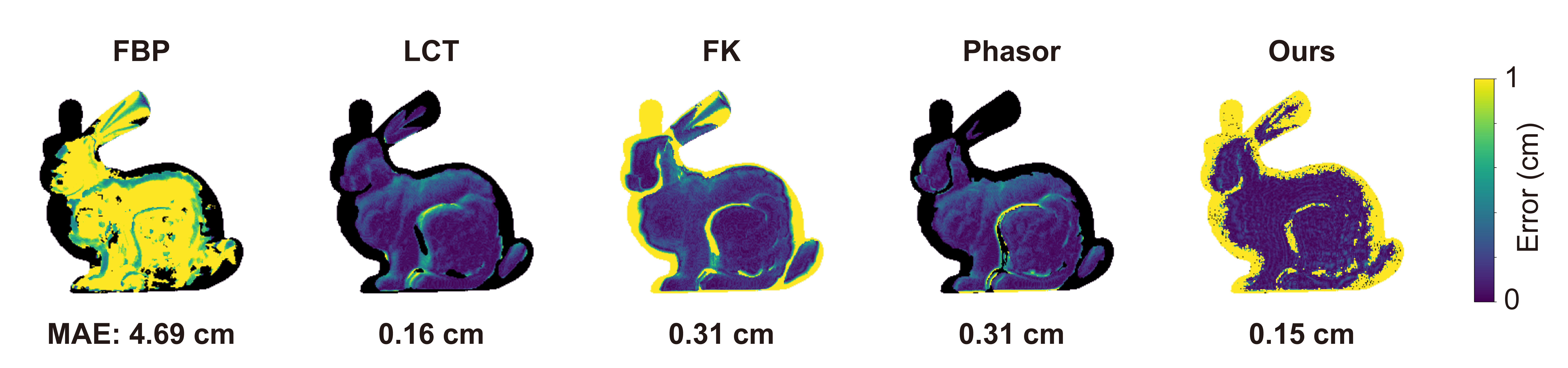}
\caption{Reconstructions of Zaragoza dataset. ``MAE" denotes the mean absolute error. The proposed method can provide depth reconstruction with smallest error.}
\label{fig: depth error}
\end{figure*}

We evaluate our algorithm using public datasets, acompare its performance with previous state-of-the-art methods, including filtered back projection (FBP) \cite{velten2012recovering}, light-cone transform (LCT) \cite{o2018confocal}, f-k migration (FK) \cite{lindell2019wave}, and phasor field (Phasor) \cite{liu2019non}.

We start with the Zaragoza dataset. The comparison of the reconstructed albedo is provided in Fig. \ref{fig: results zaragoza}, while the comparison of the reconstructed depth of the Stanford bunny is shown in Fig. \ref{fig: depth error}. The data is simulated using $256 \times 256$ scanning points within a $1 \times 1$ $\mathrm{m}^2$ region on the relay wall. Additionally, the photon travels 0.003 $\text{m}$ in each time bin.
The parameter $s$ is set to 0.025 for the USAF instance and 0.02 for the other instances.
Overall, the proposed algorithm achieves more accurate and visually faithful reconstructions compared to existing methods.

For the instances from the Zaragoza dataset, the proposed method only requires around 0.17 s to reconstruct the hidden scenes, while previous methods take over 6.39 s, more than 37 times longer than the proposed method. Additionally, the memory usage of the proposed method is less than 50 MB, which is a loose upper bound due to inaccuracies in MATLAB profiling. In contrast, previous methods require over 13.105 GB of memory, which is more than 260 times that of the proposed method. This comparison demonstrates the superior efficiency of the proposed method in both runtime and memory usage.

\begin{figure*}[t]
\centering
\includegraphics[width=1.0\textwidth]{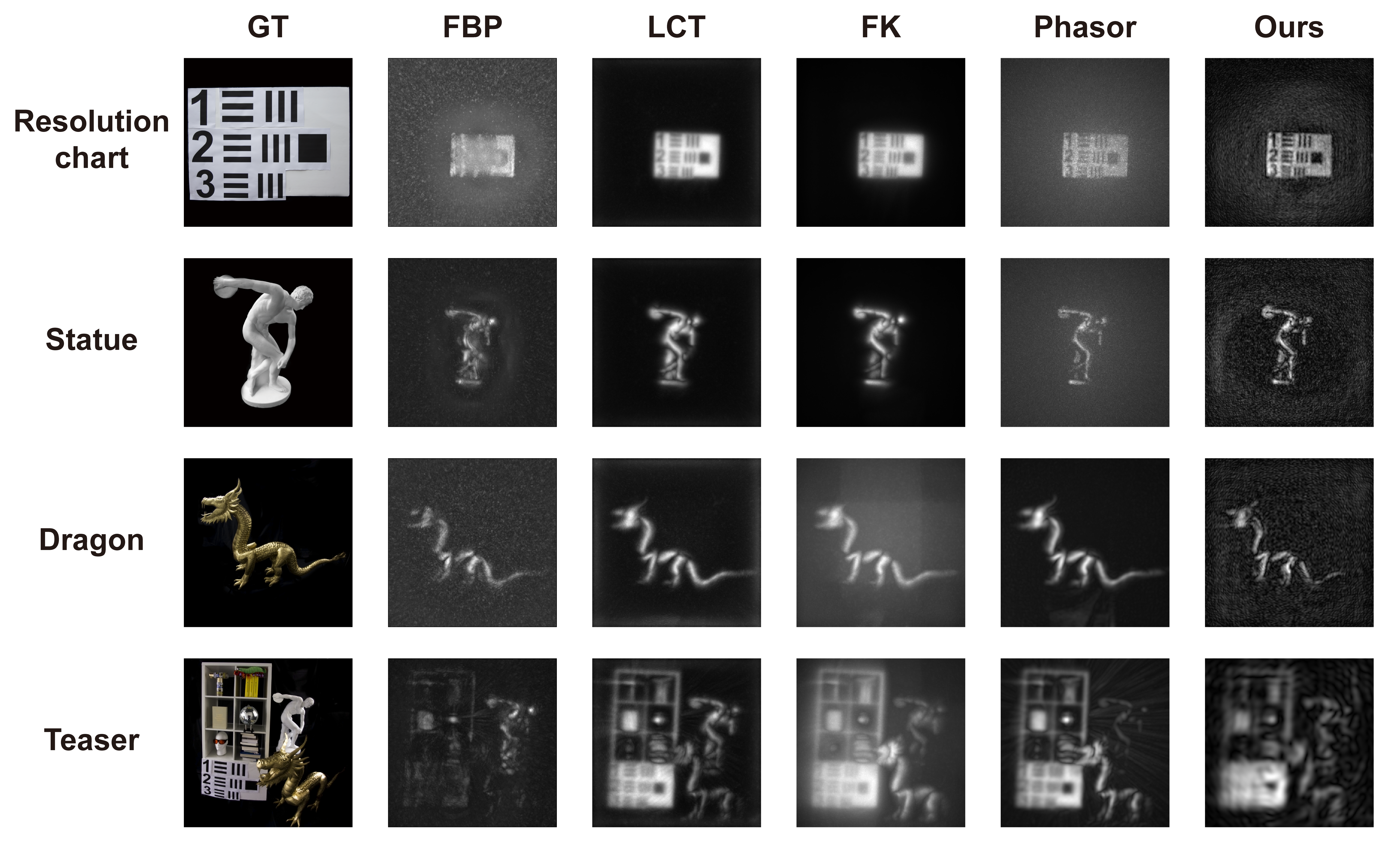}
\caption{Reconstructions of Stanford dataset. ``GT" denotes the ground truth of the hidden object. The proposed method reconstructs the hidden object with similar visual quality.}
\label{fig: results stanford}
\end{figure*}

We then test the proposed method's performance on the Stanford dataset\cite{lindell2019wave}. The reconstruction results are compared in the last two rows of Fig. \ref{fig: results stanford}. All instances have an exposure time of 180 minutes, with a measurement area of $2 \times 2$  $\mathrm{m}^2$ on the relay wall containing $512 \times 512$ scanning points. The photon travels 0.0096 m in each time bin. 
For the first two instances, the parameter $s$ is set as 0.05, while the parameter is set as 0.06 and 0.10 for the instances of dragon and teaser, respectively. 

All methods provide clear reconstructions, but the proposed method requires under 0.3 s to run and uses less than 50 MB of memory. This demonstrates the efficiency of the proposed Quasi-Fresnel transform. Furthermore, the proposed algorithm only takes the first 250 bins for reconstruction in the instance of the resolution chart, whereas LCT requires at least the first 512 time bins to achieve satisfactory results. This demonstrates that our method relies on a smaller amount of data. More detailed comparisons are provided in the Supplementary Information.

Furthermore, we validate the proposed algorithm on a dynamic scene from the Stanford dataset. The data is measured at 32 $\times$ 32 scanning points within a 2 $\times$ 2 $\mathrm{m}^2$ area on the relay wall. The human object in the hidden scene moves over time, with each frame taking 0.25 s to capture. The reconstruction results are shown in Fig. \ref{fig: results dynamic}. A video included in the Supplementary Materials further demonstrates the efficiency of the proposed method in comparison to others. While similar visual quality is achieved by all methods, the proposed method requires only 0.003 s to reconstruct one frame. If the measurement time per frame can be reduced in the future, the proposed method could enable video NLOS imaging at over 300 fps. This would significantly boost the development of practical NLOS imaging.

\begin{figure*}[t]
\centering
\includegraphics[width=0.7\textwidth]{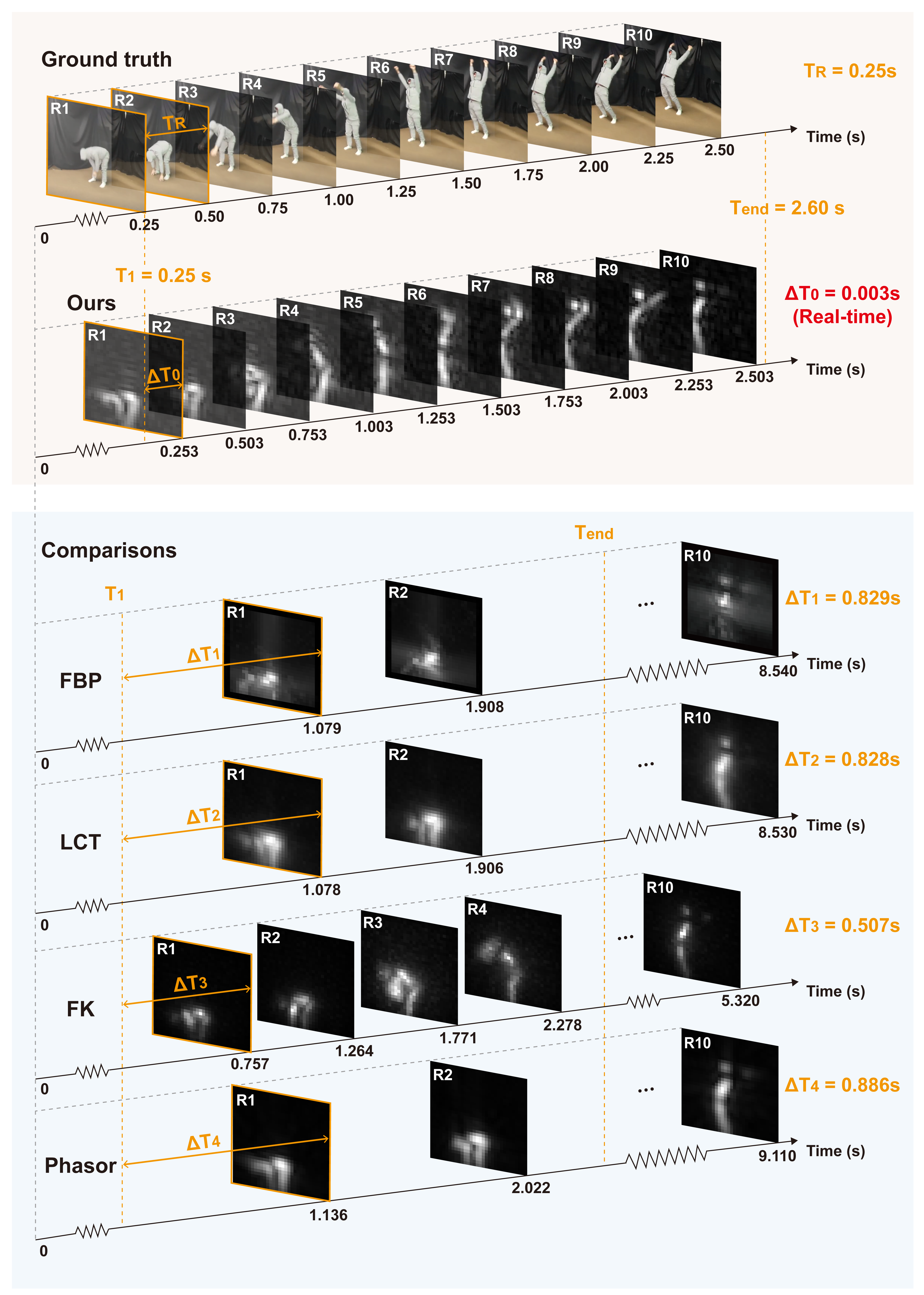}
\caption{Reconstructions of the dynamic scene. In this figure, $\text{T}_{\text{R}}$ represents the measurement time for each frame, $\text{T}_{\text{1}}$ indicates the time when the first frame is measured, and $\text{T}_{\text{end}}$ marks the end of the motion. $\Delta \text{T}_{\text{i}}, \text{ } \text{i}=0,1,2,3$ represents the runtime of different methods for reconstructing a single frame. The reconstruction time per frame using the proposed method is reduced by a factor of hundreds compared to previous methods, enabling real-time video NLOS imaging. A video is also provided in the Supplementary Materials to demonstrate the efficiency of the proposed method, as previous methods experience significantly longer delays. If the measurement time for one frame can be further reduced, video imaging at over 300 fps can be achieved with the proposed method.}
\label{fig: results dynamic}
\end{figure*}

\section*{Discussion}

Due to the two-dimensional representation, data transmission, storage, and reconstruction processes are more efficient. By using an approach similar to the Fourier domain histogram\cite{liu2020phasor}, the proposed algorithm reduces memory usage to under \textbf{5 MB} when reconstructing a measurement of size $512 \times 512$. This significantly minimizes hardware requirements. More details are provided in the Methods section.

Given these characteristics, the algorithm can be implemented on GPUs to further accelerate high-resolution reconstruction, whereas previous methods might encounter memory limitations. Our approach enables high-resolution reconstructions at speeds that were previously attainable only for coarse-grained outputs. Moreover, the proposed technique makes high-resolution NLOS imaging feasible on lower-end devices such as smartphones and embedded systems, thereby enhancing its practical applicability.

\bmhead{Differences from the phasor field method}
Although our approach shares a similar workflow and also involves a tunable parameter $s$, it is fundamentally different from the phasor field method. The phasor field technique requires a parameter analogous to a focal length. Once this parameter is fixed, a single two-dimensional deconvolution can reconstruct objects only at a specific depth, necessitating multiple reconstructions to recover scenes at varying depths. In contrast, our method does not assume that objects lie approximately on a single depth plane. The reduction in complexity stems from our parameterization of the hidden scene, allowing our model to handle objects at arbitrary depths (as illustrated in the teaser example in Fig. \ref{fig: results stanford}), regardless of the value of $s$. The influence of $s$ primarily arises from signal-to-noise ratio constraints and sampling density, requiring that it be chosen within a feasible range. 

Additionally, although Gu et al.\cite{gu2023fast} have introduced a two-dimensional intermediate representation in the phasor field method, their focus is on projecting measurements from non-planar relay surfaces onto an intermediate planar surface. This is fundamentally different from our two-dimensional representation, which is directly used for reconstructing the hidden object.

\bmhead{Further enhancement of the practicability}
Based on the established two-dimensional relationship, image processing techniques can be naturally integrated into the subprocesses of the algorithm. For example, we apply the Wiener filter to the reconstructed albedo, with enhanced results provided in the Supplementary Information.

Furthermore, the proposed method has a natural relationship with the optical diffraction theory, which can also open up interesting research avenues. Leveraging the method’s simplicity and modularity, parts or even the entire algorithm could be implemented in hardware, such as field-programmable gate arrays or optical computing platforms, to improve its efficiency further. While SPAD arrays have become a powerful tool for real-time NLOS imaging, it remains important to develop hardware for confocal measurement, as the proposed algorithm offers a promising future for this scenario. 

\section*{Conclusions}

We have proposed an efficient NLOS imaging algorithm based on a two-dimensional representation of the hidden scene, significantly reducing both memory and runtime requirements. The method maintains reconstruction quality while enabling high-resolution, real-time imaging on lightweight and embedded systems. This contribution enhances the practical deployment of NLOS imaging and opens new directions for hardware-friendly and scalable optical imaging solutions.

\section*{Materials and methods}

In this section, we describe the reconstruction algorithm based on the Quasi-Fresnel Transform, along with its complexity analysis and details of the runtime and memory usage comparison.

\subsection*{Reconstruction pipeline}

As illustrated in Fig.~\ref{fig: pipeline}, the proposed algorithm consists of the following three substeps:

\bmhead{Integration over time}The first step computes a two-dimensional function $\phi(\boldsymbol{x}, s)$ for a selected parameter $s$. According to Eq.~\eqref{eq: blur 3d}, $\phi(\boldsymbol{x}, s)$ is obtained by performing a one-dimensional weighted integration of the measurement data $\tau(\boldsymbol{x}, t)$ along the time axis $t$. 

\bmhead{Deconvolution in space} The second step computes $\psi(\boldsymbol{x}, s)$ from $\phi(\boldsymbol{x}, s)$. Eq.~\eqref{eq: relationship between functions} enables a way to directly recover $\psi$ from $\phi$. Detailed derivations of this formula are provided in Section S.2.1 of the Supplementary Information.

\bmhead{Results extraction} Finally, he albedo and depth information are recovered from $\psi(\boldsymbol{x}, s)$ according to its definition in Eq.~\eqref{eq: define psi}.

Thus, the reconstruction pipeline of the proposed algorithm is summarized in Alg. \ref{alg: simplified}. Section S.2 of the Supplementary Information provides a more detailed description of the reconstruction procedure, including strategies for selecting the parameter $s$ and enhancing the reconstruction quality.

\begin{algorithm}[t]
    \caption{The pseudocode of the proposed algorithm}
    \label{alg: simplified}
    \begin{algorithmic}[1]
        \Require the measured photon intensity $\tau(\boldsymbol{x},t)$, and the parameter $s$
        \Ensure Teo-dimensional functions $a(\boldsymbol{x})$ and $d(\boldsymbol{x})$.

        \State $  \phi(\boldsymbol{x}; s) =\frac{c^{k+1}}{2^{k+1}}\int_0^{\infty} e^{- i \frac{c^2 t^2}{16 s^2}  } t^k \tau(\boldsymbol{x}, t)  d t$

        \State $  \psi(\boldsymbol{x}; s) =  \frac{1}{16 \pi^2 s^4} \int_{\mathbb{R}^2} e^{ i \frac{\| \boldsymbol{\tilde{x}} - \boldsymbol{x}\|^2 }{4 s^2} } \phi(\boldsymbol{\tilde{x}}; s) d \boldsymbol{\tilde{x}}$

        \State $a(\boldsymbol{x}) = |\psi(\boldsymbol{x},s)|$, $d(\boldsymbol{x}) = \left(-2i s^3 \psi^{-1} \frac{\partial \psi}{\partial s}\right)^{1 / 2}$

    \end{algorithmic}
\end{algorithm}

\subsection*{Analysis of the complexity}

For a measurement with $N \times N$ scanning points and $N$ time bins, the first step of the proposed algorithm has both computational and memory complexity of $O(N^3)$ when the proposed method is executed in the same manner as previous methods, by directly loading the $N \times N \times N$ histogram.. The second step involves a two-dimensional convolution, which can be efficiently implemented using a two-dimensional FFT with a computational complexity of $O(N^2 \log N)$. According to Eq.~\eqref{eq: define psi}, the albedo and depth information can be recovered from $\psi(\boldsymbol{x}, s)$ in $O(N^2)$ time. Therefore, under this situation, the overall computaional and memory complexity are all $O(N^3)$. This is already an improvement over existing approaches. Nevertheless, the complexity of the proposed method can be further lowered with additional optimizations. Here, we will introduce two specific strategies aimed at further reducing the computational and memory demands.

A straightforward way to further reduce the memory complexity is to modify the data loading process. By leveraging the established two-dimensional mapping, the proposed algorithm requires only the aggregated measurement data to reconstruct the hidden scene. This modification enables the integration process to be performed during data loading by reading the histogram for each time bin sequentially. Consequently, although the computational complexity remains $O(N^3)$, the memory complexity can be reduced to $O(N^2)$.

Another approach to reducing both computational and memory complexity is to obtain the aggregated measurement directly from the hardware. Inspired by the Fourier domain histogram (FDH)\cite{liu2020phasor}, the aggregated measurement 
$\phi(\boldsymbol{x},s)$ can be expressed as
\begin{equation}
\label{eq: FDH ours}
    \phi(\boldsymbol{x},s) = \sum_{n=1}^N (\frac{c T_n}{2})^k e^{- \pi i \omega (\frac{c T_n}{2})^2},
\end{equation}
where $\omega = \frac{1}{4\pi s^2}$ and $k$ is $4$ for isotropic scattering or $2$ for retroreflective materials. In this way, the integration is performed during data acquisition.
Therefore, the function $\psi(\boldsymbol{x},s)$ can be obtained through a two-dimensional fast Fourier transform (FFT), whose computational complexity is $O(N^2\log N)$  and memory complexity is $O(N^2)$. 

\begin{table}
    \begin{tabular}{c|c|ccc}
        & FBP/LCT/FK/Phasor & \multicolumn{3}{c}{Ours} \\
        \cmidrule(lr){3-5}
        &  & Traditional & Loading & FDH \\
        \midrule
        Runtime & $O(N^3\log N)$  & $O(N^3)$ & $O(N^3)$ & $O(N^2\log N)$ \\
        Memory  & $O(N^3)$        & $O(N^3)$ & $O(N^2)$ & $O(N^2)$
    \end{tabular}
\caption{Performance comparison between prior state-of-the-art methods and the proposed approach. “Traditional” denotes the direct loading of the full $N \times N \times N$ histogram. “Loading” refers to sequentially loading the histogram across individual time bins. “FDH” represents the measurement obtained as described in Eq.~\ref{eq: FDH ours}. The computational complexity shown for the FBP method corresponds to its FFT-based implementation.}
    \label{tab: compare}
\end{table}

The comparison of computational and memory complexity is summarized in Tab. \ref{tab: compare}. All three versions of the proposed method outperform previous state-of-the-art algorithms. 

When reconstructing the hidden scene with $512 \times 512$ scanning points and using double-precision floating-point storage, the FDH implementation of the proposed method requires only 5 MB of memory.
Specifically, during processing, only two $\phi(\boldsymbol{x},s)$ arrays with very close $s$ values are needed (which is explained in Sec. S2 of the Supplementary Information), each using $512 \times 512 \times 8\,/\,2^{20} = 2\ \textrm{MB}$ of memory. Furthermore, a two-dimensional convolution kernel is unnecessary since the filter we use is separable, meaning only two one-dimensional filters need to be stored, requiring just tens of KB of additional memory. As a result, the total memory usage remains under 5 MB, marking a significant improvement in the efficiency of NLOS imaging.

\subsection*{Details of the runtime and memory usage comparison}

For all illustrated runtime and memory usage comparisons in Fig. \ref{fig: runtime comparison}, all five methods are implemented in MATLAB and run on the same CPU server. The server is equipped with an Intel Xeon Platinum 8358P CPU, 1 TB of memory, and runs on an Ubuntu system. For the memory usage comparison, the maximum memory usage of all methods is recorded while reconstructing the hidden scene. For the runtime comparison, each method is run five times, and the mean runtime is computed. Details of the rendered data and relevant reconstructions are provided in the Supplement Information. In addition, a video is provided as supplementary materials to further demonstrate the efficiency of the proposed algorithm.

\section*{Declarations}

\bmhead{Code availability}
The code supporting this manuscript will be available online after publication.

\bmhead{Data availability}
The data supporting this manuscript will be available online after publication.

\bmhead{Funding}

This work was supported by the National Natural Science Foundation of China (12471399, 12071244) and the Beijing Natural Science Foundation (JQ23021).

\bmhead{Competing interests}
The authors declare no competing interests.

\bmhead{Authors' contributions}
L.Q. conceived the project. Y.W. developed and implemented the Quasi-Fresnel transform. J.W. and L.Q. rendered the data. Y.W. and J.W. conducted the experiments with conceptual advice from Z.S., X.F., and L.Q. Y.W., J.W., L.X., and L.Q. composed the figures and videos. Y.W. and J.W. wrote the manuscript with contributions from all authors. All authors discussed the results and commented on the manuscript.

\bmhead{Acknowledgements}

Not applicable.

\bmhead{Supplementary Information} Supplementary Information is available in the online version of the paper.









\bibliography{sn-bibliography}

\backmatter



\end{document}